\documentclass[10pt,a5paper,headings=small,listof=totoc,headsepline,DIV=calc,BCOR=12mm,captions=tableheading,fleqn,runningheads,final]{ocmproc}

\usepackage[english,ngerman]{babel}
\usepackage{graphicx}
\usepackage[utf8]{inputenc}

\usepackage{listings}
\usepackage[font=footnotesize,labelfont=bf,format=hang]{caption}
\usepackage{color}
\usepackage[table]{xcolor}
\usepackage[standard,thmmarks]{ntheorem}
\usepackage{url}
\usepackage{amsmath}
\usepackage[sort&compress,square,numbers]{natbib}

\usepackage[T1]{fontenc}
\usepackage[sc]{mathpazo}
\usepackage[scaled]{helvet}

\newcolumntype{?}{!{\vrule width 1.1pt}}

\begin{document}
  
\mainmatter

\selectlanguage{english}  

\title{Supervised and Unsupervised Textile Classification via Near-Infrared Hyperspectral Imaging and Deep Learning}

\titlerunning{Textile Classification via Hyperspectral Imaging and Deep Learning}

\author{Maria Kainz\inst{1}, Johannes K. Krondorfer\inst{2}, Malte Jaschik\inst{1}, Maria Jernej\inst{1}, Harald Ganster\inst{1}}

\authorrunning{M. Kainz, \textit{et al.}}

\tocauthor{M. Kainz\inst{1}, J. K. Krondorfer\inst{2}, M. Jaschik\inst{1}, M. Jernej\inst{1}, H. Ganster\inst{1}}

\institute{JOANNEUM RESEARCH Forschungsgesellschaft mbH, DIGITAL,\\
Steyrergasse 17, 8010 Graz, Austria \and
Graz University of Technology, Institute of Experimental Physics \\
Petersgasse 16, 8010 Graz, Austria}

\abstract{Recycling textile fibers is critical to reducing the environmental impact of the textile industry. Hyperspectral near-infrared (NIR) imaging combined with advanced deep learning algorithms offers a promising solution for efficient fiber classification and sorting. In this study, we investigate supervised and unsupervised deep learning models and test their generalization capabilities on different textile structures. We show that optimized convolutional neural networks (CNNs) and autoencoder networks achieve robust generalization under varying conditions. These results highlight the potential of hyperspectral imaging and deep learning to advance sustainable textile recycling through accurate and robust classification.}


\maketitle

\section{Introduction}
The textile industry is a major contributor to global pollution, mainly due to its resource-intensive production processes and massive waste generation. A significant amount of textiles are discarded each year, even though they could be recycled. To counteract the negative effects, it is essential to improve textile recycling rates, as the European Union's guidelines require 100\% of textile waste to be recycled by~2025.\cite{kainz_maria:EU2021,kainz_maria:tischberger2023}

The hyperspectral near-infrared (NIR) imaging technology offers a potential solution for efficient recycling strategies. By analyzing the spectral signatures of different fibers – whether natural, artificial, or synthetic – this technology can accurately classify textiles based on their chemical composition. When combined with advanced classification algorithms, it enables textile recognition and efficient sorting of various textile types.\cite{kainz_maria:tischberger2023,kainz_maria:huang2022,kainz_maria:rodgers2009}
While standard machine learning approaches have been successfully applied to hyperspectral data~\cite{kainz_maria:plaza2009} in the past, deep learning (DL) methods typically outperform these standard methods, due to the high versatility inherent to DL models. Especially, DL models based on convolutional neural networks (CNNs), have shown strong performance in previous work, demonstrating their capacity for accurate classification and generalization.\cite{kainz_maria:wenqian2022,kainz_maria:grewal2023,kainz_maria:paoletti2019,kainz_maria:Riba2022}

In this study, we investigate the application of DL algorithms for textile classification. We explore two key use cases. First, we focus on supervised classification, where unseen textile samples are classified into one of 12 pre-defined categories \--- including pure textile fibers such as cotton, polyester, silk, wool, viscose, and nylon, cotton-polyester blends in various mixing ratios as well as cotton-elastane and linen-polyester-viscose blends. Second, we examine unsupervised classification, where only data from a single textile category are available for training. In this scenario, an autoencoder network is employed to identify normal and anomalous data based on reconstruction performance, detecting textiles that deviate from the class norm. This method proves useful in situations where not all potential textile classes are available for training, a common challenge in real-world applications.

To ensure realistic performance evaluation, we use test sets that differ from the training sets in aspects such as color, weave and yarn thickness, providing a more robust assessment of the model's out-of-sample performance and generalization. Our results demonstrate that optimized convolutional neural networks (CNNs) and autoencoder networks achieve robust generalization under varying conditions, showcasing the potential of combining hyperspectral imaging with deep learning for automated textile sorting.

\section{Textile data, Acquisition and Preprocessing}\label{kainz_maria:sec:data}
\begin{figure}[b!]
    \includegraphics[width=0.33\textwidth]{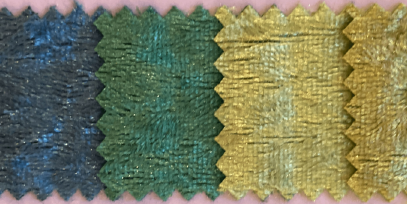}%
    \includegraphics[width=0.33\textwidth]{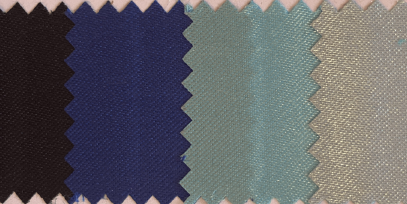}%
    \includegraphics[width=0.33\textwidth]{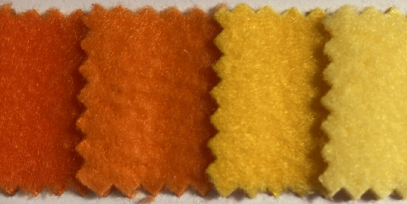}%
    \caption{Comparison of polyester structure types (Panne Velvet, Satin, Fleece).}
    \label{kainz_maria:fig:textile_samples}
\end{figure}

\paragraph{Textile categorization: }
First, we describe the textile data recorded in this study and explain how they are categorized and summarized in Table~\ref{kainz_maria:tab:datasets}. Primarily, the textiles are categorized by their \textit{fiber type}, which represents the basic raw material of a textile. Fibers are divided into natural, artificial and synthetic fibers. Natural fibers either come from plants, such as cotton or linen, or from animals, such as wool or silk. Artificial fibers are derived from materials of natural origin. However, the production of yarn takes place through chemical transformation (e.g. viscose from cellulose). Unlike the previous ones, synthetic fibers (e.g. polyester) are produced by chemical synthesis.
Combinations of different fiber types are called blended fibers.
Here, we aim to classify textiles according to their fiber type, to facilitate the recycling of fibers.
In addition to the fiber type, the \textit{structure type} or \textit{production type}, provides information about the processing of the textile fibers into textile end products. Fibers are spun into yarns, which are then processed into woven or knitted fabrics, which can differ in yarn properties, thread density, fabric type and other manufacturing techniques. There are also non-woven fabrics that are made directly from fibers and not from yarn. 
Fiber type and structure type together result in the \textit{textile type} (\textit{fiber type} + \textit{structure type} = \textit{textile type}). Textile samples within a textile type, i.e. a row in Table~\ref{kainz_maria:tab:datasets}, differ purely in their coloring. This is also illustrated in Figure~\ref{kainz_maria:fig:textile_samples} for different types of polyester.
\begin{table}[ht!]
\centering
\rowcolors{2}{gray!10}{white}
\begin{scriptsize}
    \centering
    \caption{Specification of textile data and assignment of objects to different data sets. The objects are differentiated by their fiber- and structure-type. Each line corresponds to one textile-type, that can entail different colors. The first column shows abbreviations used throughout the manuscript. D1 - D6 indicate the assignment of textiles to different data sets, that are studied in this paper. D1 is used for training, validation, and testing of the supervised classification model. D2 and D3 are used to evaluate the generalization and out-of-sample performance of the supervised model, differing in color and structure, respectively. Training of the unsupervised classification is performed on D4, and the performance on pure textiles and blends is tested via D5 and D6, respectively.}\vspace{4pt}
    \label{kainz_maria:tab:datasets}
    \begin{tabular}{r?l|c?l|c?c|c|c?c|c|c}
    \multicolumn{1}{c?}{\textbf{}} & \multicolumn{2}{c?}{\textbf{Fiber}} & \multicolumn{2}{c?}{\textbf{Structure}} & \multicolumn{3}{c?}{\textbf{Sup. Cl.}} & \multicolumn{3}{c}{\textbf{Unsup. Cl.}}  \\
    \textbf{Textile} & \textbf{Type} & \textbf{\%}& \textbf{Type} & \textbf{Nr.} & \textbf{D1} & \textbf{D2} & \textbf{D3} & \textbf{D4} & \textbf{D5} & \textbf{D6} \\
    \hline
    C1 & Cotton (plant-based) & 100 & Flag Fabric & 1 &  45 & 15 & & 60 & & \\
    C2 & Cotton & 100 & Twill & 2 & & & 29 & & &29\\
    C3 & Cotton & 100 & Cord & 3 & & & 22 & & &22\\
    C4 & Cotton & 100 & Canvas & 4 & & & 32 & & &32\\
    C5 & Cotton & 100 & Cretonne & 5 & & &  37 & & &37 \\
    P1 & Polyester (synthetic) & 100 & Panne Velvet & 1 &  37 & 12 & & &49 & \\
    P2 & Polyester & 100 & Satin & 2 & & &  22 & & 22& \\
    P3 & Polyester & 100 & Fleece & 3 & & &  47 & & 47& \\
    S1 & Silk (animal-based) & 100 & - & 1 & 1 & 1 & & &2 & \\
    S2 & Silk & 100 & - & 2 & & &  1 & & 1 & \\
    L1 & Linen (plant-based) & 100 & - & 1 & 3 & 1 & & &4 & \\
    N1 & Nylon (synthetic) & 100 & Organza & 1 & 11 & 4 & & &15 &\\
    W1 & Wool (animal-based) & 100 & Walkloden & 1 &  41 & 14 & & &55 & \\
    V1 & Viscose (artificial) & 100 & Voile & 1 & 28 & 9 & & &37& \\
    VLP1 & Viscose/Linen/Polyester & 40/30/30 & - & 1 & 25 & 8 & & &33 &\\
    CP1 9:1 & Cotton/Polyester & 90/10 & Frottee & 1 & 19 & 6 & & &25& \\
    CP1 8:2 & Cotton/Polyester & 80/20 & Sweatshirt & 1 & 24 & 8 & & &32& \\
    CP1 7:3 & Cotton/Polyester & 70/30 & Sweatshirt & 1 & 38 & 12 & & &50 & \\
    CE1 & Cotton/Elastane & 95/5 & Jersey & 1 & 62 & 21 & & &83 & \\
    CE2 & Cotton/Elastane & 95/5 &  French Terry &2  & & & 48 & & 48& \\
    \hline
    \end{tabular}
\end{scriptsize}
\end{table}

\paragraph{Hyperspectral imaging: } 
The textile samples were analyzed on a conveyor belt with a pushbroom hyperspectral camera (\textit{ImSpector~N17E} spectograph \cite{kainz_maria:ImSpector} and a \textit{PhotonFocus} camera \cite{kainz_maria:PhotonF}). This setup was used to examine the spectrum between 990~nm and 1700~nm (400~channels).

\paragraph{Dark textiles:} 
It was found that black and dark-grey samples of textile type C5 and dark-blue samples of textile type CP1~9:1 absorb almost all radiation, thus providing no information about the fiber type in the recorded reflection spectrum. These samples are excluded from the dataset. This phenomenon is also mentioned in Ref.~\cite{kainz_maria:tischberger2023}, where it is attributed to the complete absorption of radiation by carbon pigments. For all other dark-colored samples, this phenomenon was not observed.

\paragraph{Preprocessing:} To mitigate the influence of the light source's spectral signature and the background reflection pattern, the measured intensities are calibrated by subtracting the dark reference and normalizing with respect to the white reference. Furthermore, we apply Standard Normal Variate (SNV) to remove baseline drift, followed by mean smoothing (i.e. averaging 5x5 pixel into one spectrum) to reduce noise, and apply a first derivative Savitzky-Golay-Filter to emphasize subtle spectral features, ensuring improved data quality for subsequent analysis. The combination of SNV+Mean+SG showed the best performance during hyperparameter optimization for both investigated scenarios.

\section{Supervised Classification of Hyperspectral Textile Data}
\paragraph{Datasets:} For the supervised classification task, three datasets were prepared to evaluate the model's performance and generalization capabilities, as summarized in Table~\ref{kainz_maria:tab:datasets}. We perform stratified splitting of D1 (1300 spectra per class) into training~(60\%), validation~(20\%), and test~(20\%) subsets. D2 consists of spectral data from textile samples exhibiting visible color variations compared to D1. And D3 is composed of spectral data from textiles differing in structural properties such as yarn thickness, thread density, and weave pattern. This dataset design enables a robust evaluation of model generalization under different conditions of variability.

\paragraph{Model:} The final model employs a one-dimensional convolutional neural network optimized for hyperspectral data. The model is obtained by variation of different hyperparameters, such as number of layers, activation functions, normalization, and preprocessing. The final architecture employs the (SNV+Mean+SG) preprocessing discussed in Section~\ref{kainz_maria:sec:data}, and an input layer size of 400, corresponding to the number of channels in the hyperspectral dataset. This is followed by two 1d-convolutional layers with ReLU activation function and a kernel size of~5. The number of filters is set to 20 and 32 respectively. Then a fully connected dense layer is used, with a batch normalization layer and a dropout layer with rate 0.5, followed by a ReLU activation and a 12~class softmax output layer.

\paragraph{Training:} The network is trained using the categorical cross-entropy loss function and optimized with the Adam optimizer~\cite{kainz_maria:adam} with initial learning rate (LR) of 0.001, and a mini-batch size of 128. A LR-scheduler is used to reduce the LR by a factor of 0.2 when the validation loss failed to improve for five consecutive epochs. Early stopping with a patience of seven epochs is employed to prevent overfitting.

\begin{figure}[htb!]
\centering
    \includegraphics[width=\textwidth]{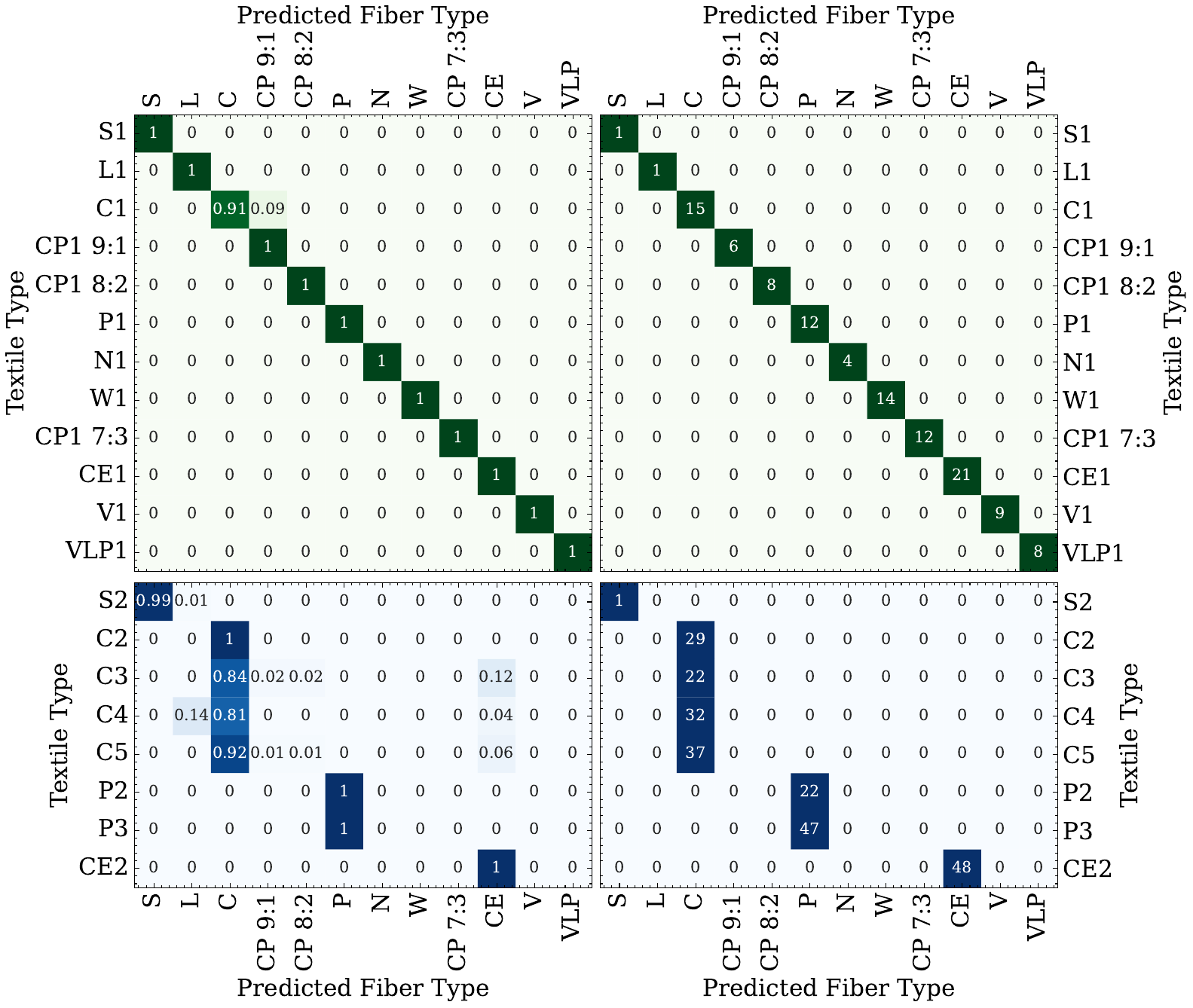}
    \caption{Confusion matrices for supervised classification of pixel-based (left) and object-based (right) analysis on D2 (top) and~D3 (bottom). On the test set of D1 all data are correctly classified. Also on D2 and~D3 exceptional performance can be observed, with slight misclassification rates for cotton blends or structurally different cotton fibers. With respect to an object-based accuracy, however, all textiles are classified correctly, despite the different structure and coloring.}
    \label{kainz_maria:fig:confusion_matrix_supervised}
\end{figure}

\paragraph{Results:} To assess the model's performance, we compute pixel-based accuracy (i.e. the fraction of correctly classified pixels across all samples) and object-based accuracy (determined by majority voting of classified pixels assigned to an object), and display the corresponding confusion matrices in Figure~\ref{kainz_maria:fig:confusion_matrix_supervised}. On the test set of D1 all samples are classified correctly. Also on D2~and~D3 exceptional performance can be observed, only with slight misclassification rates for some cotton blends and structurally different cotton fibers for a pixel-based analysis. With respect to an object-based analysis, however, perfect classification is observed across all data sets, showing the exceptional generalization and out-of-sample performance of this comparably simple model.%

\section{Unsupervised Classification Using Autoencoders}
\paragraph{Datasets:} For the unsupervised classification task, datasets are structured to evaluate the autoencoder’s ability to filter target textiles. As test case, we choose cotton as target textile. We train the model on D4 (1300 spectra) of Table~\ref{kainz_maria:tab:datasets}, with a stratified split into training (60\%), validation (20\%), and test (20\%) subsets. We test the detection capabilities in D5, consisting of non-cotton textiles and different cotton blends, and D6, comprised of cotton textiles, differing in structural properties, to assess robustness.

\paragraph{Model:} The final autoencoder model, designed for target textile detection in hyperspectral data, is optimized through systematic hyperparameter variations, including the number of layers, activation functions, and units per layer. It employs preprocessing steps (SNV+Mean+SG) discussed in Section~\ref{kainz_maria:sec:data}. The input and output layers are fixed at size~400, i.e. the number of spectral channels. The encoder compresses the input into a 20-dimensional latent space using two fully connected layers with 100 units and ReLU activations. The decoder mirrors the encoder, consisting of two layers with 100 units, reconstructing the input via a linear activation in the final layer.

\paragraph{Training:} The autoencoder is trained to minimize the mean squared error~(MSE) between input and reconstructed data. The Adam optimizer~\cite{kainz_maria:adam} is used for training with an initial LR of 0.001, and a mini-batch size of~16. A LR-scheduler is used to reduce the LR by a factor of 0.5 if the MSE does not improve over five consecutive epochs. Early stopping, monitoring the MSE of the validation set, is implemented with a patience of seven epochs to prevent overfitting. 

\begin{figure}[htb!]
\centering
\includegraphics[width=\textwidth]{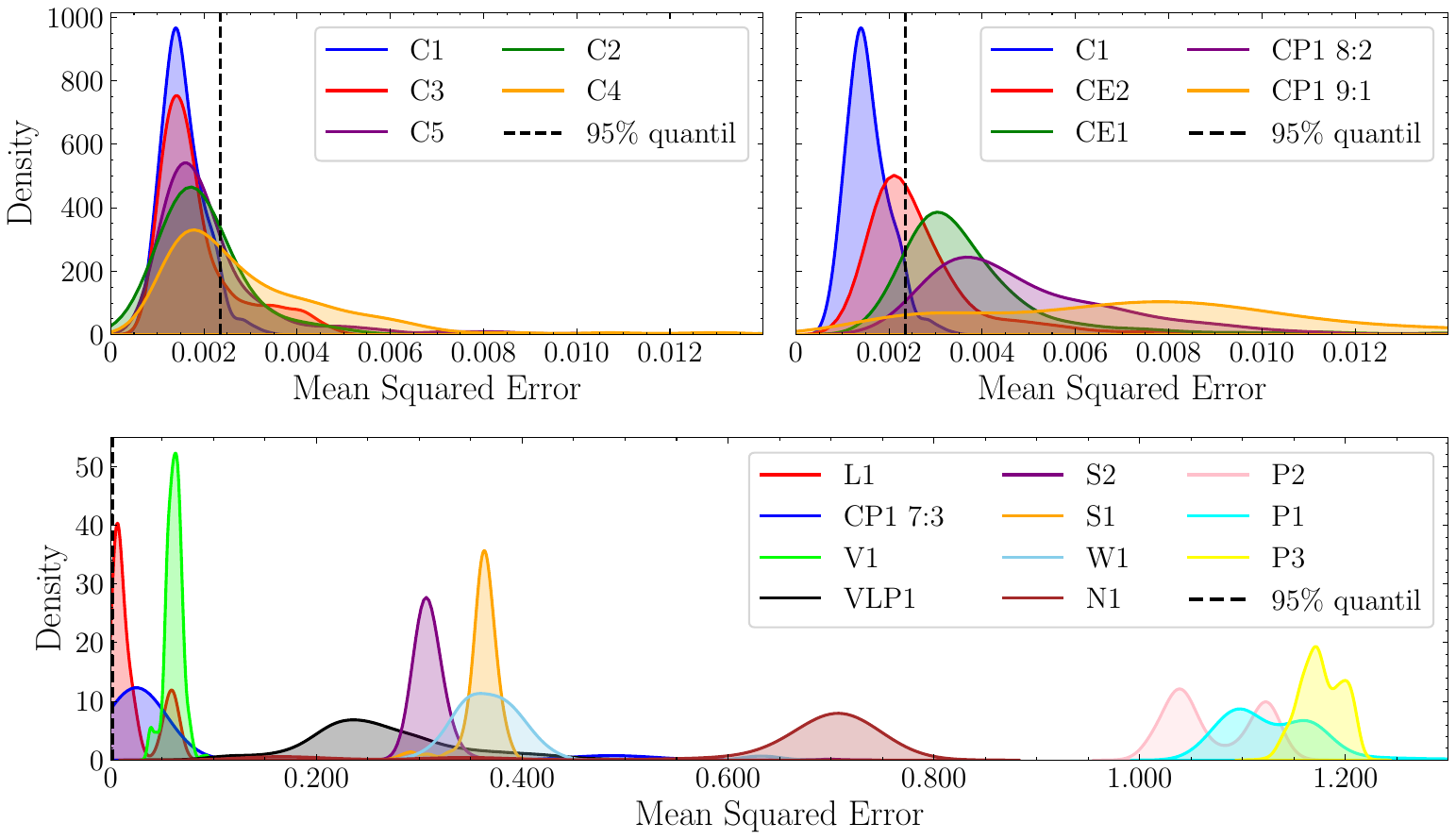}
\caption{Distribution of reconstruction error (RE) for different textiles. The decision threshold is based on the 95\% quantile of the RE distribution over D4. Pure textiles from D5 can be well distinguished and yield 100\% recognition (bottom), whereas cotton blends are much harder to distinguish and only reduced recognition can be observed (upper left). The RE distribution of different cotton types from D6 is much wider than on the training set and only moderate generalization is observed, leading to reduced accuracy. The corresponding pixel- and object-based accuracies are given in Table~\ref{kainz_maria:tab:autoencoder_results}.}
    \label{kainz_maria:fig:mse distribution}
\end{figure}

\paragraph{Results:} To assess the model's performance, we compute the reconstruction error (RE) on the different data sets. The classification is based on the observed RE value and the 95\% quantile of the RE distribution over D4. Samples with RE lower (higher) than this threshold value are classified as cotton (non-cotton). The results are summarized in Figure~\ref{kainz_maria:fig:mse distribution}, where the distribution of the RE over D4,~D5~and~D6 is shown, and Table~\ref{kainz_maria:tab:autoencoder_results}, where pixel- and object-based accuracies of the textiles are listed. The autoencoder effectively identifies non-cotton textiles based on reconstruction error, with significant separation observed for synthetic fibers (e.g., polyester and nylon). Cotton-polyester blends exhibited intermediate RE values, making them more challenging to classify correctly. Structurally different cotton types from D6 show a wider RE distribution than on the training set and only moderate generalization is observed, leading to reduced accuracy.
\begin{table}[htb!]
\rowcolors{2}{gray!10}{white}
\centering
\begin{footnotesize}
    \centering
    \caption{Pixel- and object-based accuracies (in~\%) of unsupervised textile classification.}\vspace{4pt}
    \label{kainz_maria:tab:autoencoder_results}
    \begin{tabular}{c|c|c|c|c|c|c|c|c|c|c|c|c|c|c|c|c|c|c|c}
     & P1 & S1 & L1 & N1 & W1 & V1 & \shortstack{CP1\\9:1} & \shortstack{CP1\\8:2} & \shortstack{CP1\\7:3} & CE1 & VLP1 & P2 & P3& S2 & CE2 & C2 & C3 & C4 & C5\\
     \hline
     px & 100 & 100 & 100 & 100 & 100 & 100 & 90 & 98 & 100 & 90 & 100 & 100 & 100 & 100 & 41 & 76 & 56 & 87 & 77 \\
     obj & 100 & 100 & 100 & 100 & 100 & 100 & 92 & 100 & 100 & 100 & 100 & 100 & 100 & 100 & 42 & 86 & 69 & 86 & 83  \\
    \end{tabular}
\end{footnotesize}
\end{table}

\section{Conclusion and outlook}
In this study, we investigated supervised and unsupervised deep learning models and their generalization capabilities. For this purpose, we tested the performance of the developed models on data sets that differ from the training sets in aspects such as color and textile structure.

The supervised classification approach showed exceptional performance for hyperspectral textile data, achieving high in-sample and out-of-sample accuracy. However, its success depends on the availability of a representative and diverse dataset for different fiber types, which can be challenging, due to limited availability of all relevant fibers. 

In contrast, the unsupervised approach using autoencoders showed lower performance but is less restrictive in terms of dataset requirements, as only one fiber type is needed for training. This makes it a promising alternative for scenarios where comprehensive labeled datasets are not available. Due to reduced performance on blends, this technique is better suited for pre-sorting of different fibers. 

Overall, the results highlight that even relatively simple models, when suitably optimized and combined with preprocessing, can achieve robust classification and generalization performance. This provides a solid foundation for further advancements in automated textile analysis and recycling applications.

\section*{Acknowledgements}
Funded by the Federal Ministry for Climate Action, Environment, Energy, Mobility, Innovation, and Technology (BMK).\newpage

\begingroup
\renewcommand{\bibsection}{\section*{References}}
\normalsize
\setlength{\bibsep}{2pt}
	\renewcommand{\bibname}{References}
	\bibliographystyle{IEEEtran}

\endgroup

\end{document}